\documentclass{article}
\usepackage{spconf,amsmath,graphicx,hyperref}
\usepackage{booktabs,multirow,subcaption}

\title{What You Feel Is Not What They See: On Predicting Self-Reported Emotion from Third-Party Observer Labels}

\name{Yara El-Tawil*\thanks{*Denotes equal contribution.} \qquad Aneesha Sampath*\footnotemark[1] \qquad Emily Mower Provost }
\address{University of Michigan, Ann Arbor, Michigan, USA}

\begin{document}
%
\maketitle
\begin{abstract}
Self-reported emotion labels capture internal experience, while third-party labels reflect external perception. These perspectives often diverge, limiting the applicability of third-party–trained models to self-report contexts. This gap is critical in mental health, where accurate self-report modeling is essential for guiding intervention. We present the first cross-corpus evaluation of third-party-trained models on self-reports. We find activation unpredictable (CCC $\approx$ 0) and valence moderately predictable (CCC $\approx$ 0.3). Crucially, when content is personally significant to the speaker, models achieve high performance for valence (CCC $\approx$ 0.6–0.8). Our findings point to personal significance as a key pathway for aligning external perception with internal experience and underscore the challenge of self-report activation modeling.
\end{abstract}
\section{Introduction}
\label{sec:intro}

Self-reported emotion labels capture an individual's own evaluation of their emotion state, whereas third-party (perceived) labels reflect how an outside observer interprets the expression \cite{zhang2016automatic}. Crucially, these perspectives may diverge, as the producer's intentions may not align with the perceiver's interpretation \cite{zhang2016automatic, li-etal-2025-third, busso2008expression}. This gap is significant in contexts such as mental health, where modeling of self-reports is essential for assessing symptoms and guiding intervention \cite{gideon2019emotion}. Yet, self-report labels are both rare and difficult to predict, while large-scale corpora typically provide third-party labels only.

Prior work has compared joint and separate modeling of self-reports and third-party labels within a single dataset, showing that the two label types diverge and are difficult to reconcile \cite{zhang2016automatic, busso2008expression}. Moreover, inter-rater agreement between speaker self-reports and observer ratings is consistently lower than agreement among observers themselves, reflecting systematic misalignment between internal experience and external perception \cite{busso2008iemocap, jaiswal2019muse}. These findings highlight both the necessity and difficulty of modeling self-reported emotion. Open questions remain: (1) can models trained on third-party labels generalize across corpora to predict self-reports? (2) do the dimensional emotion attributes of activation (calm–excited) and valence (negative–positive) \cite{russell1979affective} differ in their predictability? and (3) can speaker-centered factors determine when alignment between perspectives is possible?

We center our analysis on activation and valence. Activation is conveyed primarily through paralinguistics (audio), whereas valence is more closely tied to linguistics (text) \cite{calvo2010affect}. We also examine emotion predictability through speaker-specific factors. Appraisal theory posits that emotions arise from evaluations of novelty, pleasantness, goal relevance, and personal significance \cite{ellsworth2002appraisal}. Personal significance refers to the extent to which a statement reflects content that is relevant and consequential to the speaker's internal state. We assess whether personal significance explains when third-party models succeed in capturing self-reports. To evaluate, we leverage both self-supervised speech and language transformers \cite{wagner2023dawn, sampath2025efficient, qin2023bert} and large language models (LLMs) \cite{liu2024emollms}.

Our contributions are three-fold: (1) We present the first cross-corpus analysis of third-party-trained models evaluated on self-reports. We find that performance drops from third-party to self-report evaluations across datasets. (2) We provide an attribute-specific comparison, showing that activation diverges sharply (Concordance Correlation Coefficient (CCC) $\approx$ 0) while valence is moderately predictable (CCC $\approx$ 0.3). (3) We show that speaker-significant content enables generalization, with third-party–trained models reaching valence CCC $\approx$ 0.6–0.8 on estimated high-significance samples.

Our findings suggest that activation self-reports remain largely unpredictable across training setups and modality. Valence self-reports can be modeled effectively when accounting for the significance of the content to the speaker. This has direct implications for model design and dataset development.

\section{Research Questions}

\emph{RQ1:} \textit{Can models trained on third-party labels generalize across corpora to predict self-reports?} Emotion recognition has primarily focused on third-party training and evaluation \cite{wagner2023dawn, lotfian2017buildingpodcast, sampath2025efficient}. Hence, we evaluate models trained on third-party labels, both third-party and self-report labels, and off-the-shelf LLMs on cross-corpus self-report prediction. We hypothesize that models will exhibit poor performance given the within-corpus challenges in prior work \cite{zhang2016automatic}. 

\emph{RQ2:} \textit{Are activation and valence equally predictable, or does one generalize more reliably than the other?} Prior work shows that activation and valence diverge in predictability, often by modality \cite{calvo2010affect, wagner2023dawn}, and less often in the context of self-report \cite{truong2009arousal}. Here, we hypothesize that valence will generalize more effectively than activation, as prior work found higher agreement between self-report and third-party for valence than for activation \cite{truong2009arousal}. We evaluate models separately for self-reported activation and valence across modalities. 

\emph{RQ3:} \textit{Do speaker-centered factors determine when alignment between perspectives is possible?} Appraisal theory posits that emotions arise from cognitive evaluations of events, such as personal significance \cite{ellsworth2002appraisal}. Speakers' internal states may be presented more outwardly when the content carries personal significance.
We hypothesize that self-reports will align with third-party labels when the content is significant. We group samples by an estimated significance score and evaluate performance across groups.

\section{Datasets}
\label{sec:datasets}

\textbf{MSP-Podcast} contains non-acted English conversational speech from podcasts \cite{lotfian2017buildingpodcast}. We use version 1.11, which includes 151,654 segments from 2,172 speakers across 237 hours of audio and manual transcriptions. We train on the `Train' split and evaluate on the `Test1' split, and select the model with the best performance on the `Development' split.

\textbf{Multimodal Stressed Emotion (MuSE)} contains non-acted audiovisual English monologues \cite{jaiswal2019muse}. We generate transcripts with word-level timestamps using Whisper (openai/ whisper-large-v2) \cite{radford2023robust}. Speakers provided \textit{momentary} activation and valence ratings after each monologue, which we apply at both the monologue and segment level. 
MuSE includes 273 monologues (10 hours, 2,648 segments; $\sim$9.7 segments/monologue), with segments averaging 28.4 words and 12.5s, and monologues averaging 275.3 words and 120.9s. We measure the Concordance Correlation Coefficient (CCC) ~~\cite{lawrence1989concordance} between self-report monologue-level labels and third-party segment-level labels by assigning the monologue-level to each segment within the monologue. We find no agreement for activation (CCC = –.039) and moderate agreement for valence (CCC = .299). When averaging the third-party segment-level labels over the monologue and comparing them to monologue-level labels, we find no agreement for activation (-.179) and improved agreement for valence (.350).



\textbf{IEMOCAP} contains acted audiovisual English conversations with manual transcripts \cite{busso2008iemocap}. Of 10,039 segments (12 hours), we retain 2,411 with \textit{retrospective} segment-level self-report labels provided by speakers after re-watching their recordings.
The 2,411 segments average 11.4 words and 4.3s. 
The CCC between third-party and self-report labels is 0.561 (activation) and 0.781 (valence). Correlations between third-party and self-report labels are higher in IEMOCAP than in MuSE, likely due to IEMOCAP's \textit{retrospective} self-reports versus MuSE's \textit{momentary} self-reports.
This is likely why IEMOCAP self-reports are more aligned with third-party compared to those of MuSE. CCC is consistently higher for valence than activation in both datasets. We scale labels into the $[0,1]$ range as in \cite{wagner2023dawn, sampath2025efficient}. We evaluate on MuSE and IEMOCAP. To the best of our knowledge, they are the only audio datasets with \textit{dimensional} self-report emotion labels.

\section{Method}
\label{sec:method}

\label{ssec:configs}

\begin{figure}[t]
    \centering
    \includegraphics[width=0.46\textwidth]{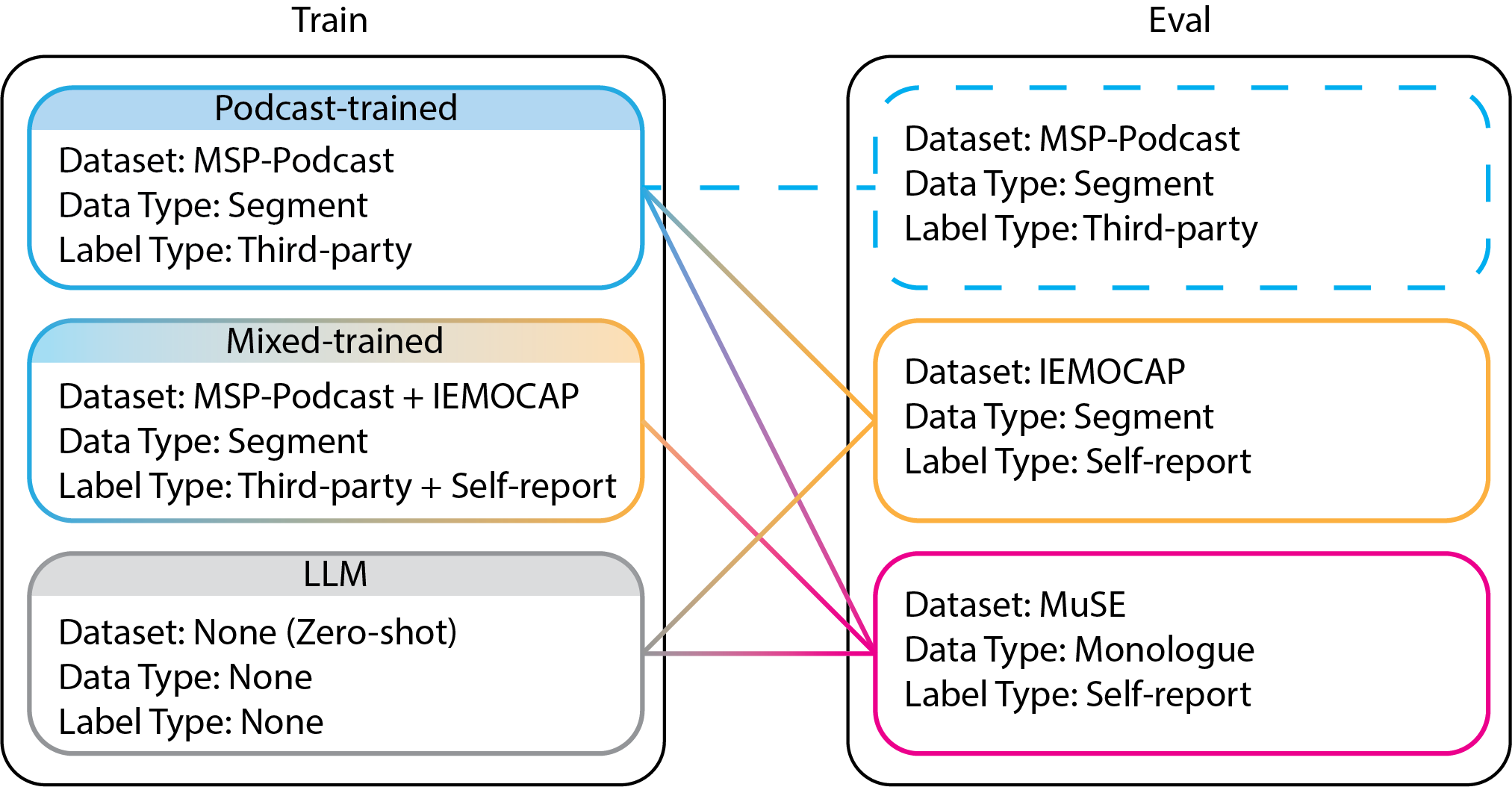}
    \caption{
    Left: model training setups. 
    Right: evaluation setups.}
    \label{fig:configs}
\end{figure}


\textbf{Training.} We train under three settings (left column, Figure~\ref{fig:configs}). In the first, we train models on MSP-Podcast third-party segment-level labels (Podcast-trained). In the second, we further train these models on the full set of IEMOCAP self-reports (Mixed-trained). We do not train on MuSE due to the absence of segment-level self-report labels. In the third, we use off-the-shelf LLMs.


\textbf{Third-Party Evaluation.} We evaluate Podcast-trained models on third-party segment-level labels using MSP-Podcast Test1 split, IEMOCAP, and MuSE. This contextualizes self-report performance with third-party performance.

\textbf{Self-Report Evaluation.} We evaluate Podcast-trained models and LLMs on MuSE (monologue) and IEMOCAP (segment) self-reports. We evaluate Mixed-trained models on MuSE self-reports only (we use all IEMOCAP self-report data in finetuning). We use two aggregation strategies: \textit{Seg-Mono}, applying the monologue-level self-report label to each segment in the monologue (2,648 evaluations), and  \textit{Avg-Segs}, averaging segment-level predictions across a monologue (273  evaluations) (see Figure~\ref{fig:configs}, right column). We explore segment-level aggregation strategies to align segment-level training inputs with monologue-level evaluation.

\textbf{Self-report Evaluation with Speaker-Centered Factors}. We group samples by their estimated personal significance to the speaker and evaluate self-report performance. We use gpt-oss-20B \cite{agarwal2025gpt} to assign each sample a significance score (1–5), motivated by prior work showing GPT's effectiveness for appraisal evaluation \cite{ruder2025assessing}. 
We design a prompt that assesses significance based on critical factors (e.g., real vs. hypothetical, personal vs. external). We manually review a random 10\% of GPT labels to identify obvious errors. Future work should incorporate expert annotations.

\subsection{Architecture}
For audio, we use WavLM \cite{chen2022wavlm}, a self-supervised encoder. Training follows prior work \cite{wagner2023dawn, sampath2025efficient}: we freeze the CNN feature extractor, mean-pool the final transformer hidden states, apply dropout $(p=0.2)$, and finetune the last three transformer layers, with separate heads for activation and valence. 

For language, we use both self-supervised language encoders and LLMs. We use RoBERTa-Longformer \cite{beltagy2020longformer} and ModernBERT \cite{warner2024smarter}. They have long maximum token lengths to better capture full monologues from MuSE. 
For both models, we apply dropout $(p=0.2)$ to the CLS token, pass it through a shared linear layer, apply dropout $(p=0.2)$, and finetune all transformer layers, with separate heads for activation and valence. For LLMs, we use state-of-the-art open source models gpt-oss-20B \cite{agarwal2025gpt} and Qwen3-32B \cite{yang2025qwen3}. Model checkpoints and prompts are listed on GitHub\footnote{github.com/chailab-umich/self\_report\_prompts/}.

\subsection{Model Training and System Configuration}
For the first two training settings, we compute the loss separately for activation and valence, and then average them. We train with AdamW and CCC loss \cite{wagner2023dawn}. We use a fixed learning rate of 1e-5 for WavLM, and 2e-5 for Longformer and ModernBERT. We use batch size 32. We train in FP16, as in \cite{sampath2025efficient}. For `Podcast-trained', we train for five epochs and select the model with the best performance on the MSP-Podcast `Development' set. For `Mixed-train', we further finetune those models on IEMOCAP self-report labels for 2 epochs. 

We load LLMs in BF16 with default hyperparameters. We allow three retries per sample. We do not perform finetuning or few-shot prompting; instead, we evaluate them zero-shot. 

We run experiments on an HPC cluster with NVIDIA A40 GPUs (48GB). We train WavLM, Longformer, and ModernBERT on a single A40. LLM inference uses two A40s.

\section{Results and Analysis}
\label{sec:results}
We evaluate each setting with five random seeds (five inference runs for LLMs). We report mean CCC ± standard deviation over the dataset. We assess significance with a one-way ANOVA followed by Bonferroni-corrected t-tests. We use paired t-tests for Tables ~\ref{table:tp_results} and ~\ref{tab:sr_results}. Each model is compared against the best-performing model (bolded) within each row. We use unpaired t-tests for personal significance groups (Figure~\ref{fig:critical_appraisal_figs}). We assert significance when $p < .05$. 

\begin{table}[t]
\centering
\small
\setlength{\tabcolsep}{4pt}
\begin{tabular}{l l c c c}
\toprule
\textbf{Dataset} & \textbf{Dim} & \textbf{WavLM} & \textbf{Longformer} & \textbf{ModernBERT} \\
\midrule

\multirow{2}{*}{MSP-Podcast} 
 & Act & \textbf{0.65 ± 0.00} & 0.32 ± 0.01$\dagger$ & 0.31 ± 0.00$\dagger$ \\
 & Val & \textbf{0.60 ± 0.00} & 0.55 ± 0.01$\dagger$ & 0.54 ± 0.02$\dagger$ \\
\midrule

\multirow{2}{*}{IEMOCAP} 
 & Act & \textbf{0.61 ± 0.03} & 0.34 ± 0.01$\dagger$ & 0.38 ± 0.02$\dagger$ \\
 & Val & \textbf{0.45 ± 0.01} & 0.41 ± 0.01 & 0.39 ± 0.03 \\
\midrule

\multirow{2}{*}{MuSE} 
 & Act & \textbf{0.48 ± 0.05} & 0.11 ± 0.02$\dagger$ & 0.11 ± 0.01$\dagger$ \\
 & Val & 0.42 ± 0.05 & \textbf{0.62 ± 0.03}* & 0.64 ± 0.01* \\
\bottomrule
\end{tabular}
\caption{Third-party results for Podcast-trained models. Mean CCC ± std for activation (Act) and valence (Val). $\dagger$ = significantly worse, * = significantly better than WavLM.}
\label{table:tp_results}
\end{table}

\begin{table*}[t]
\centering
\footnotesize
\setlength{\tabcolsep}{5pt}
\begin{tabular}{@{}ll@{\hspace{6pt}}*{3}{c}@{\hspace{6pt}}*{3}{c}@{\hspace{6pt}}*{2}{c}@{}}
\toprule
\multirow{2}{*}{\textbf{Dataset}} & \multirow{2}{*}{\textbf{Dim}} 
& \multicolumn{3}{c}{\textbf{Third-party Training (Podcast)}} 
& \multicolumn{3}{c}{\textbf{Mixed Training (Podcast + IEMOCAP)}} 
& \multicolumn{2}{c}{\textbf{Zero-shot}} \\
\cmidrule(lr){3-5} \cmidrule(lr){6-8} \cmidrule(lr){9-10}
& & \textbf{WavLM} & \textbf{Longformer} & \textbf{ModernBERT} 
& \textbf{WavLM} & \textbf{Longformer} & \textbf{ModernBERT} 
& \textbf{GPT-20B} & \textbf{Qwen3-32B} \\
\midrule

\multirow{2}{*}{MuSE} 
& Act & 0.029 ± 0.008$\dagger$ & 0.079 ± 0.028 & 0.045 ± 0.018$\dagger$ & 0.005 ± 0.001$\dagger$ & \textbf{0.131 ± 0.017} & 0.119 ± 0.046 & 0.039 ± 0.030$\dagger$ & N/A \\
& Val & 0.258 ± 0.020$\dagger$ & 0.317 ± 0.017 & 0.305 ± 0.019 & 0.040 ± 0.008$\dagger$ & \textbf{0.342 ± 0.011} & 0.268 ± 0.031 & 0.336 ± 0.010 & N/A \\

\multirow{2}{*}{IEMOCAP} 
& Act & \textbf{0.534 ± 0.033} & 0.335 ± 0.009$\dagger$ & 0.348 ± 0.011$\dagger$ & -- & -- & -- & 0.242 ± 0.008$\dagger$ & 0.169 ± 0.006$\dagger$ \\
& Val & \textbf{0.423 ± 0.016} & 0.393 ± 0.008$\dagger$ & 0.368 ± 0.010$\dagger$ & -- & -- & -- & 0.399 ± 0.010 & 0.359 ± 0.005$\dagger$ \\

\bottomrule
\end{tabular}
\caption{Self-report CCC on IEMOCAP and MuSE. Columns: training setting (Fig~\ref{fig:configs}); rows: dataset; Values: mean ± std activation (Act) and valence (Val). N/A: invalid output on $\geq$ 15 samples. $\dagger$ = significantly worse than bolded value in each row.}
\label{tab:sr_results}
\end{table*}

\begin{table*}[t]
\centering
\footnotesize
\setlength{\tabcolsep}{3pt}
\begin{tabular}{@{}ll@{\hspace{6pt}}*{3}{c}@{\hspace{6pt}}*{3}{c}@{\hspace{6pt}}*{2}{c}@{}}
\toprule
\multirow{2}{*}{\textbf{Dataset}} & \multirow{2}{*}{\textbf{Dim}} 
& \multicolumn{3}{c}{\textbf{Third-party Training (Podcast)}} 
& \multicolumn{3}{c}{\textbf{Mixed Training (Podcast + IEMOCAP)}} 
& \multicolumn{2}{c}{\textbf{Zero-shot}} \\
\cmidrule(lr){3-5} \cmidrule(lr){6-8} \cmidrule(lr){9-10}
& & \textbf{WavLM} & \textbf{Longformer} & \textbf{ModernBERT} 
& \textbf{WavLM} & \textbf{Longformer} & \textbf{ModernBERT} 
& \textbf{GPT-20B} & \textbf{Qwen3-32B} \\
\midrule

\multirow{2}{*}{MuSE (Seg-Mono)} 
& Act & -0.049 ± 0.009 & 0.051 ± 0.009 & 0.042 ± 0.018 
       & -0.002 ± 0.001 & 0.061 ± 0.007 & 0.062 ± 0.010 
       & -0.011 ± 0.016 & N/A \\ 
& Val & 0.110 ± 0.016 & 0.191 ± 0.007 & 0.199 ± 0.012 
       & 0.013 ± 0.002 & 0.202 ± 0.009 & 0.200 ± 0.008 
       & \textbf{0.259 ± 0.023} & N/A \\ 
\addlinespace[0.2em]

\multirow{2}{*}{MuSE (Avg-Segs)} 
& Act & -0.040 ± 0.010 & 0.076 ± 0.010 & 0.075 ± 0.022 
       & -0.001 ± 0.001 & 0.081 ± 0.008 & 0.085 ± 0.013 
       & -0.016 ± 0.018 & -0.095 ± 0.018 \\
& Val & 0.115 ± 0.020 & 0.281 ± 0.017 & 0.281 ± 0.019 
       & 0.012 ± 0.002 & \textbf{0.353 ± 0.009} & 0.349 ± 0.016 
       & 0.250 ± 0.021 & 0.261 ± 0.018 \\
\bottomrule
\end{tabular}
 \caption{Aggregation analysis on MuSE. Columns show training settings (Figure~\ref{fig:configs}); rows report activation (Act) and valence (Val). Strategies: (i) Seg-Mono (ii) Avg-Segs (see Section \ref{sec:method}). Results are mean ± std of CCC.}
\label{tab:muse_aggregation}
\end{table*}

\subsection{RQ1: Models Trained on Third Party Labels}
\label{ssec:rq1}


Cross-corpus prediction is strong for third-party labels. Audio is significantly better for activation across datasets, and language is significantly better for valence on MuSE (Table~\ref{table:tp_results}).

Self-report results for MuSE (monologues) and IEMOCAP (segments) are in Table ~\ref{tab:sr_results}. Seg-Mono and Avg-Segs aggregation are in Table~\ref{tab:muse_aggregation}. \textbf{Cross-corpus performance drops across all models compared to third-party predictions.} Performance drops substantially compared to third-party prediction. IEMOCAP exhibits more stability, which is likely due to stronger self-third-party label alignment (Section~\ref{sec:datasets}).

Overall, cross-corpus prediction to self-reports is unreliable for MuSE, but comparatively reliable for IEMOCAP. 

\subsection{RQ2: Generalizability to Activation and Valence}
\label{ssec:rq2}


We compare third-party and self-report performance on activation and valence (significance via unpaired t-tests).

\emph{\textbf{MuSE}.} Activation self-report CCCs are near-zero across all models, even when third-party activation CCC was strong (e.g., WavLM CCC = .48 drops to CCC = .029 (significant) for monologue input, and CCC = -.04 (significant) for Avg-Segs, Tables ~\ref{tab:sr_results} and ~\ref{tab:muse_aggregation}). Valence is modestly predictable (CCC = .258-.342; best: Mixed-train Longformer; Table~\ref{tab:sr_results}), with the exception of Mixed-train WavLM, which achieves CCC = .04, suggesting that the additional IEMOCAP self-report finetuning step hinders cross-corpus generalization for self-report valence prediction from audio. This is likely due to the acted nature of IEMOCAP, which does not generalize to the naturalistic monologues of MuSE. Aside from Mixed-train WavLM, the divergence between activation and valence generalization is persistent across modalities and training setups. 

\emph{\textbf{IEMOCAP}.} Podcast-trained WavLM shows a larger drop for activation than valence when moving from third-party to self-report (A: -.07 (significant); V: -.03 (significant)),
again indicating relative generalizability of valence under self-report even for audio (Table~\ref{tab:sr_results}). More broadly, all models perform better on IEMOCAP than on MuSE, consistent with IEMOCAP's stronger self-third-party alignment (Section~\ref{sec:datasets}).


Overall, \textbf{self-reported activation is less generalizable across corpora}. Although self-report and third-party labels in IEMOCAP correlate moderately (CCC = .561, Section~\ref{sec:datasets}), models achieve weaker activation performance, suggesting that self-reported activation reflects less observable internal cues. The non-acted nature of MuSE likely compounds this difficulty. In contrast, \textbf{valence shows more consistent cross-corpus self-report generalization} across modalities.

\subsection{RQ3: Speaker-Centered Factors and Alignment}
\label{ssec:rq3}
\begin{figure}[t]
    \centering
    \begin{subfigure}[b]{0.44\textwidth}
        \centering
        \includegraphics[width=\linewidth]{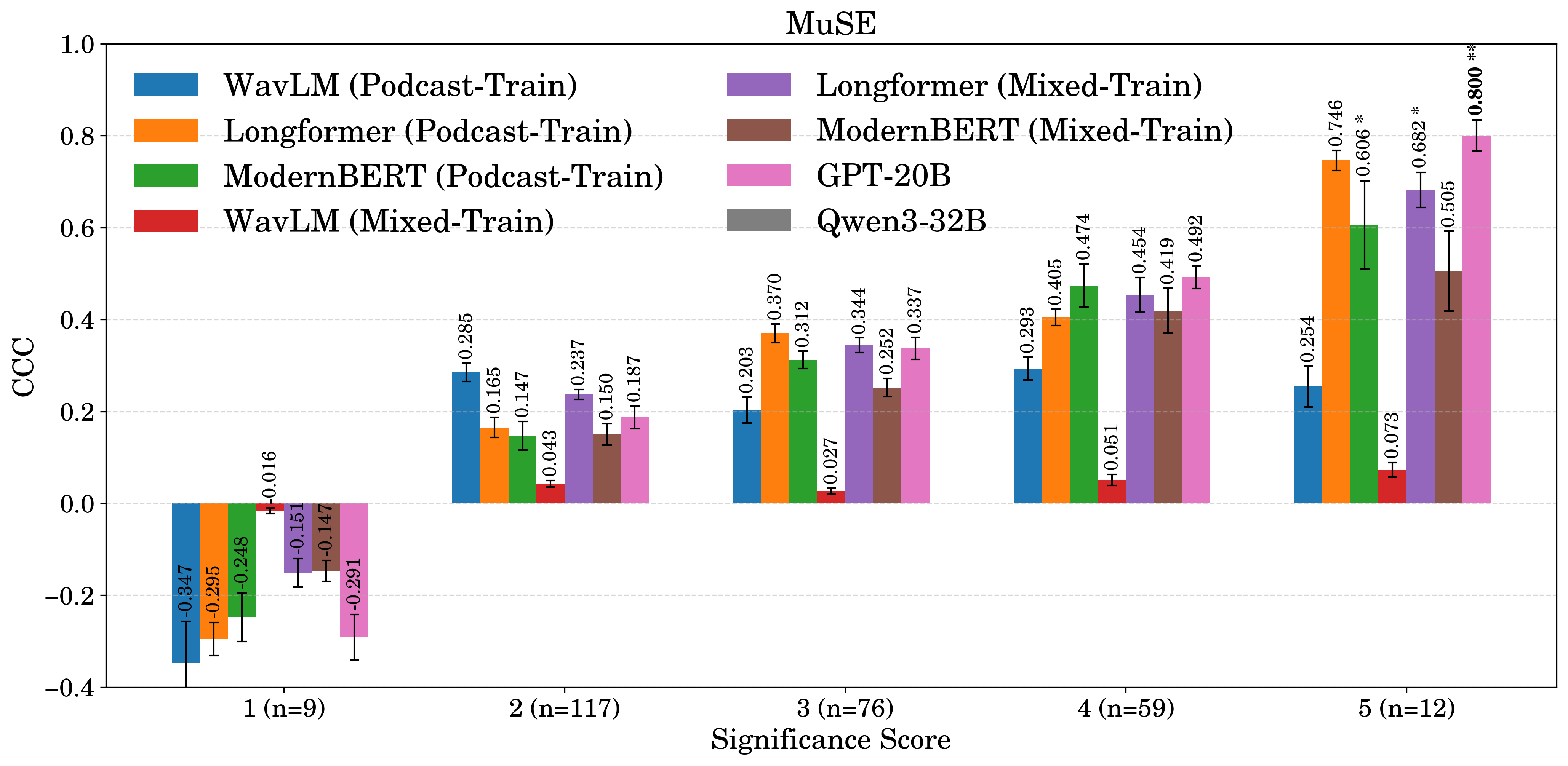}
        \label{fig:muse_critical}
    \end{subfigure}
    \begin{subfigure}[b]{0.44\textwidth}
        \centering
        \includegraphics[width=\linewidth]{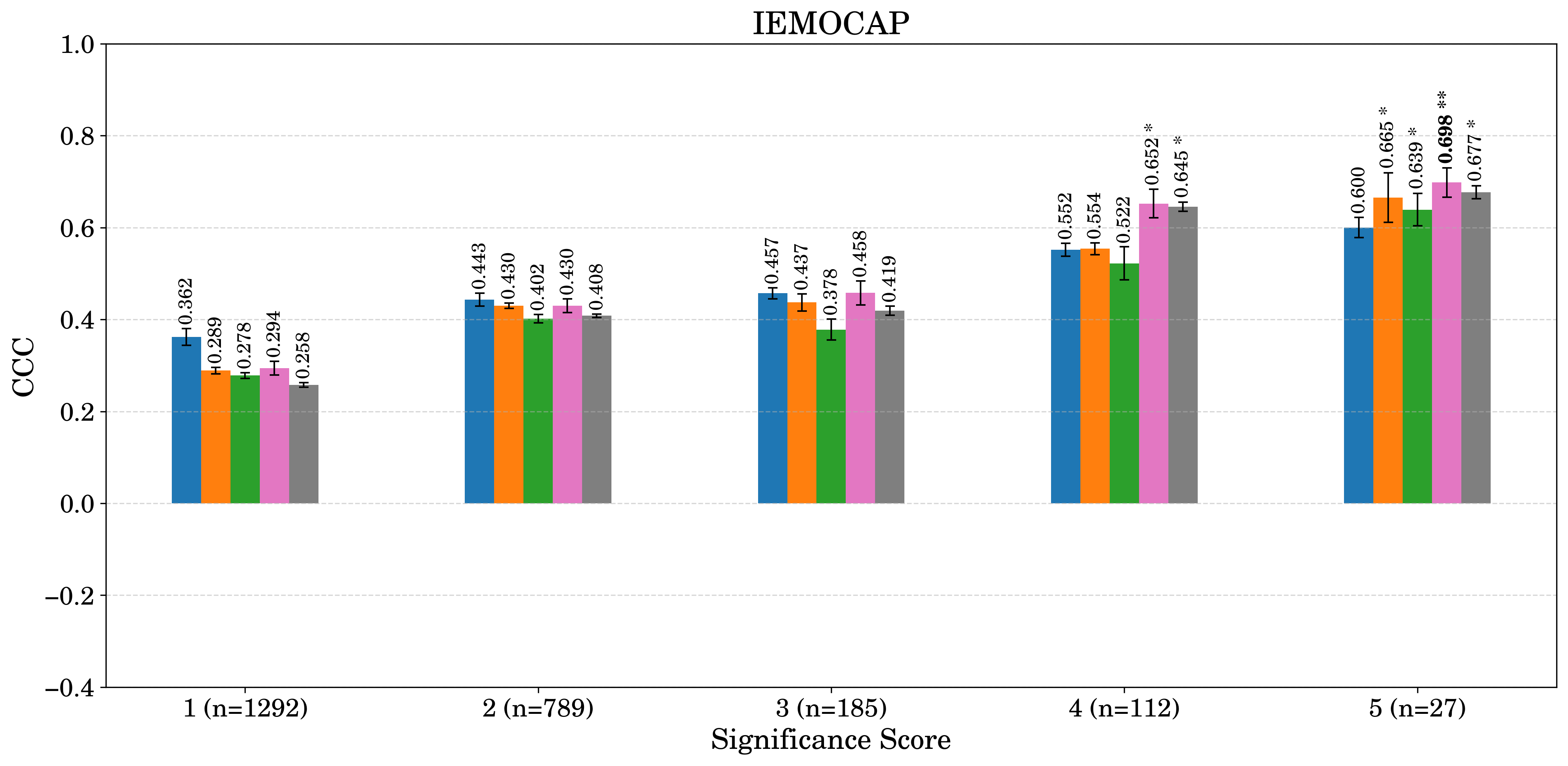}
        \label{fig:iemocap_critical}
    \end{subfigure}
    
    \caption{Valence by significance score. ** = best; * = not significant vs. best.}
    \label{fig:critical_appraisal_figs}
\end{figure}



We evaluate model performance on self-report labels at different levels of estimated personal significance (Figure \ref{fig:critical_appraisal_figs}). We report on valence only, as activation remains unreliable (Section~\ref{ssec:rq2}). Although estimated significance is not necessarily independent of valence, their empirical correlation is weak (Pearson r=0.01 on MuSE and r=-0.14 on IEMOCAP), suggesting that significance captures information beyond valence intensity. \textbf{Estimated significance level is the clearest indicator of when self-reports can be predicted from third-party models}. On MuSE, gpt-oss-20B achieves CCC = 0.8 at significance level 5 and 0.49 at level 4. On IEMOCAP, it reaches 0.7 and 0.65 at levels 5 and 4, respectively. gpt-oss-20B at level 5 significantly outperforms all models on MuSE except Longformer (Mixed-train) and ModernBERT (Podcast-train) at level 5. In contrast, at low significance (1–2), CCCs fall near zero for MuSE and $\approx$ .25 for IEMOCAP.


\section{Conclusion}
\label{sec:conclusion}
In this paper, we present a cross-corpus analysis of audio and language models to examine when third-party emotion models generalize to self-reports. We evaluated three training setups: Podcast-trained, Mixed-trained, and LLMs. Self-report activation proved largely unpredictable (CCC $\approx 0$), while valence showed modest alignment (CCC $\approx .3$). Crucially, when stratified by personal significance, third-party models achieved strong correlations with self-reported valence in personally significant samples (CCC $\approx$ 0.6-0.8). This highlights personal significance as a key factor for self-report modeling.

Across setups, Mixed-training on IEMOCAP self-report hindered generalization to MuSE for WavLM in particular, while Podcast-trained models and LLMs generalized more effectively. Within significance groups, language models consistently outperform WavLM on self-reported valence. Future work will integrate expert annotations and analyze the divergence in self-reported activation versus valence.
\\\textbf{Acknowledgments:} This material is based in part upon work supported by the National Science Foundation (NSF IIS-RI 2230172), NSF CSGrad4US (2313998), and National Institutes of Health (NIH R01MH130411).


\bibliographystyle{IEEEtran}
\bibliography{strings,refs}
\end{document}